\documentclass[journal]{IEEEtran}
\IEEEoverridecommandlockouts
\usepackage[vietnamese]{babel}
\usepackage{cite}
\usepackage{amsmath}
\usepackage{amssymb,amsfonts}
\usepackage{algorithmic}
\usepackage[pdftex]{graphicx}
\usepackage{textcomp}
\usepackage{subfigure}
\usepackage{multirow}
\usepackage{xcolor}
\usepackage{enumerate}
\def\BibTeX{{\rm B\kern-.05em{\sc i\kern-.025em b}\kern-.08em
    T\kern-.1667em\lower.7ex\hbox{E}\kern-.125emX}}

\ifCLASSINFOpdf
\else
\fi

\begin{document}
\title{Data-driven Smart Ponzi Scheme Detection}

%\author{Kai~Lei,~\IEEEmembership{Member,~IEEE,}
%        Weijing~Wu,~\IEEEmembership{Fellow,~OSA,}
%        Yuzhi~Liang,~\IEEEmembership{Life~Fellow,~IEEE}% <-this % stops a space
%\thanks{M. Shell was with the Department
%of Electrical and Computer Engineering, Georgia Institute of Technology, Atlanta,

\author{Yuzhi~Liang, Weijing~Wu, Kai~Lei*, and Feiyang~Wang
\IEEEcompsocitemizethanks{\IEEEcompsocthanksitem Yuzhi Liang is with ICNLab, Peking University, Shenzhen, China; Kai Lei, Weijing Wu, and Feiyang Wang are with ICNLab, Peking University, Shenzhen, China , and Peng Cheng Laboratory, Shenzhen, China. Kai Lei is the corresponding author, email: leik@pkusz.edu.cn}
}

\maketitle

\begin{abstract}
A smart Ponzi scheme is a new form of economic crime that uses Ethereum smart contract account and cryptocurrency to implement Ponzi scheme. The smart Ponzi scheme has harmed the interests of many investors, but researches on smart Ponzi scheme detection is still very limited. The existing smart Ponzi scheme detection methods have the problems of requiring many human resources in feature engineering and poor model portability. To solve these problems, we propose a data-driven smart Ponzi scheme detection system in this paper. The system uses dynamic graph embedding technology to automatically learn the representation of an account based on multi-source and multi-modal data related to account transactions. Compared with traditional methods, the proposed system requires very limited human-computer interaction. To the best of our knowledge, this is the first work to implement smart Ponzi scheme detection through dynamic graph embedding. Experimental results show that this method is significantly better than the existing smart Ponzi scheme detection methods.
\end{abstract}

\begin{IEEEkeywords}
Blockchain Ponzi scheme detection, data-driven method
\end{IEEEkeywords}

% \fi
\IEEEpeerreviewmaketitle

\section{Introduction}
\begin{figure}[ht]
  \centering
  \includegraphics[width=0.35\textwidth]{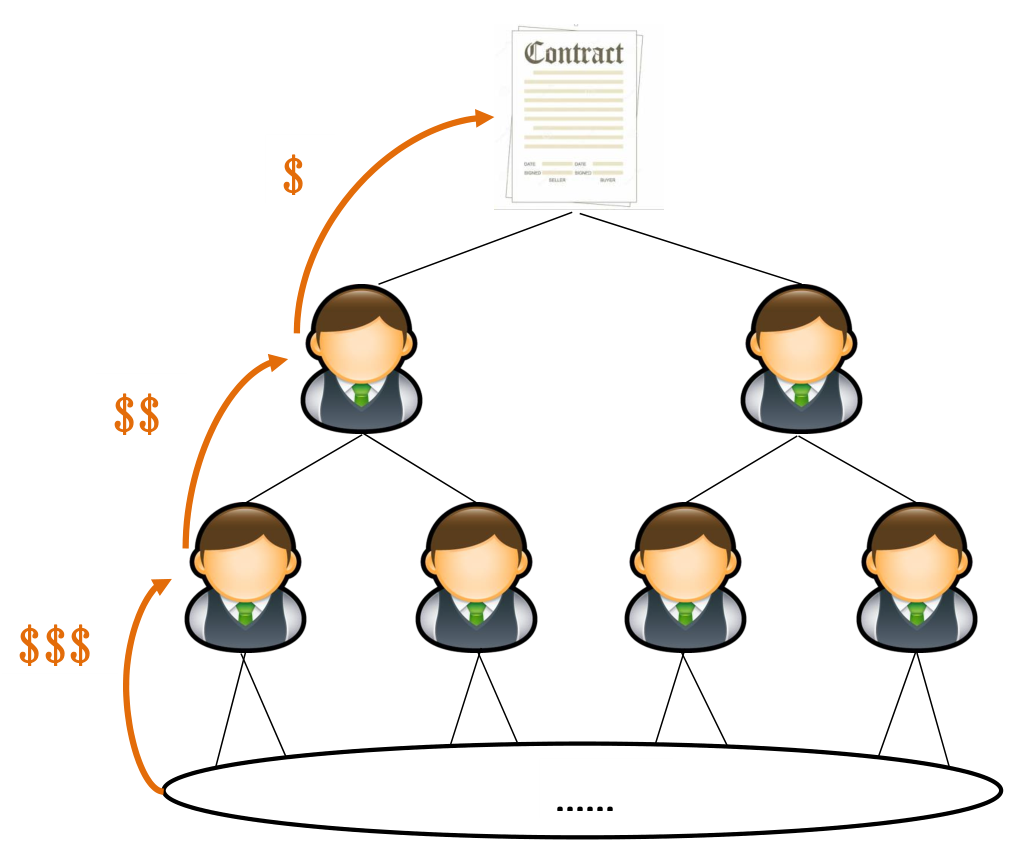}
\caption{An example of a Ponzi scheme. The top-level is the initiator of the scheme, and the investment of users at each level is compensated by users of the next layer. Ponzi schemes require a constant flow of funds from new investors. When it is difficult to recruit new investments, or a large number of investors demand financial returns, the Ponzi scheme will inevitably collapse.}\label{ponzi_example}
\end{figure}

A Ponzi scheme is an investment scheme that uses the funds contributed by new investors to pay off the returns of existing investors  (Fig. \ref{ponzi_example}). Recently, the prosperity of blockchain has given birth to a new form of Ponzi scheme, that is, the Ponzi scheme based on Ethereum. Specifically, criminals write the code for implementing the Ponzi scheme in the smart contract of the Ethereum account and realize the Ponzi scheme through the automatic execution of the smart contract \cite{elwell2013bitcoin, swan2015blockchain, zheng2016blockchain}. This new form of Ponzi scheme is called the smart Ponzi scheme. Ethereum is an excellent environment for the implementation of smart Ponzi scheme for the following reasons: 1)  The account on Ethereum is anonymous, so users cannot identify the smart Ponzi scheme account by checking the profile of the account creator; 2) The smart contract is public, immutable and self-executing, which makes it easy for investors to relax their vigilance and mistakenly believe that their investment is safe. According to the research in \cite{vasek2014empirical}, the loss of the smart Ponzi scheme to investors exceeded 7 million U.S. dollars from September 2013 to September 2014. 

It is essential to identify the accounts implementing the smart Ponzi scheme efficiently. Compared to passively waiting for the victim to report, proactive detection usually detects the problematic account faster, allowing victims have more time to take action to reduce losses. However, research on smart Ponzi scheme detection is still very limited. The existing smart Ponzi scheme detection methods can be broadly divided into detection methods based on source code inspection \cite{atzei2017survey,chen2017under} and detection methods based on feature engineering and machine learning \cite{Bartoletti2018Data,Jung2019DataME,Chenweili2018,FARRUGIA2020113318}. The detection method based on source code inspection detects the smart Ponzi scheme by manually checking the source code on the smart contract. The problem with this method is that code checking is cumbersome and requires a lot of human resources. The method based on feature engineering and machine learning represents an account on Ethereum through a set of designed features and then inputs the account representation into the machine learning model to determine whether the account is a smart Ponzi scheme account. This method usually requires a lot of professional knowledge to design account features, and it is difficult to represent the account with a limited number of features precisely. Due to the lack of good representation, advanced machine learning techniques do not perform well in feature-based smart Ponzi scheme detection. According to the experiments in  \cite{Jung2019DataME}, in the feature-based smart Ponzi pattern detection, the performance of using neural networks as classifiers is even worse than that of using decision trees as classifiers. In addition, feature-based methods may have poor portability, and features suitable for one Ethereum environment may not work properly in another Ethereum environment.

From our investigation, Ponzi scheme contracts share a few patterns, and most new Ponzi scheme contracts are obtained by modifying existing Ponzi scheme contracts. In general, the smart Ponzi scheme can be divided into four schemes, namely the array-based pyramid scheme, the tree-based pyramid scheme, the switching scheme, and the waterfall scheme \cite{Bartoletti2017Dissecting}. If we build a transaction network with Ethereum accounts as nodes and Ethereum transactions as edges, the account nodes that implement the Ponzi scheme will form a special structure in the transaction network. On the other hand, according to the existing research, we know that the temporal information and code in the smart contract are also important information for smart Ponzi scheme detection \cite{Chenweili2018}. Different transaction sequences can reflect different transaction purposes, and the operation code on the smart contract reveals the control logic of the contract account. Therefore, we can use different information sources to observe whether the target account is implementing a smart Ponzi scheme from multiple views, which is conducive to obtaining correct results for detection.

In this paper, we design a novel data-driven smart Ponzi scheme detection system named DSPSD. DSPSD integrates the structural information of each node in the transaction network, the dynamic information (i.e., the formation process of the transaction network), and the operation code of the smart contract into a low-dimensional continuous vector through dynamic node embedding. The embedding has the following characteristics. First, it can retain the structural information of the transaction network. Second, It is interactive account-aware, that is, the account has different embeddings when the account is transacting with different accounts; Thrid, it preserves the formation process of the  transaction network by tracking the trading history of each account. Finally, it contains the operation code information for each account. The contributions of this work are as follows.

\begin{enumerate}
\item We propose a data-driven smart Ponzi scheme detection system DSPSD. DSPSD can be regarded as a function that directly predicts whether the account is implementing a Ponzi scheme based on the account's opcode and transaction data. The system can automatically learn the representation of the account based on the input data, so very limited human interaction is required. Compared with the method based on feature engineering, DSPSD saves the labor cost of feature design, and at the same time has better detection effect and adaptability to the environment.
\item A method is designed to project multi-source account transaction-related information into a low-dimensional continuous vector. This method uses the vector generated based on structural information of the transaction network, the control logic of the account, and the dynamic changes of the transaction network as the representation of the account. Compared with generating node representation based on a single data source, using multi-source data can observe accounts from multiple views and generate a more comprehensive account representation.

%According to the smart Ponzi scheme patterns, we design a method to integrate information about an Ethereum account from different sources into a low-dimensional continuous vector. For each account, its corresponding vector preserves the related transaction network structure, account control logic and dynamic changes of the transaction network.
\item We have conducted extensive experiments on a large-scale dataset, and the performance of DSPSD is significantly better than the existing smart Ponzi scheme detection methods.
\end{enumerate}
\section{Background and Related Work}
\subsection{Ethereum}
Ethereum is a Blockchain-based distributed computing platform and operating system. Ethereum implements an execution environment on the Blockchain called Ethereum Virtual Machine (EVM), which can excute code of arbitrary algorithmic complexity\cite{hirai2017defining}.  Users can use existing programming languages such as Python and Javascript to create applications that run on EVM. Then, Ethereum can serve as a platform for many different types of decentralized Blockchain applications, including but not limited to cryptocurrencies.

We introduce some terms about Ethereum as follows.

\textbf{Ethereum account}: Ethererum uses an account-based system, and the state of Ethernet consists of accounts in Ethereum and transactions between accounts. Ethereum has two types of accounts, namely contract accounts and externally owned accounts (EOA) \cite{buterin2014next}\cite{dannen2017introducing}. Contract accounts are controlled by the smart contracts associated with the accounts. When a contract account receives a message, its contract code is activated. On the other hand, externally owned accounts have no smart contracts associated with them.

\textbf{Smart contract}: Smart contracts are public, self-executing code that runs when certain trigger conditions are met \cite{atzei2016survey,bogner2016decentralised,szabo1996smart}. For example, it can be a function that sends a message to a specific account when the account balance reaches a predefined value. A smart contract reflects the control logic of the corresponding contract account.

\textbf{Opcode}: Before uploading to the Blockchain, the source code in the smart contract will be compiled into an Ethereum-specific binary format called EVM bytecode. The EVM bytecode in a smart contract consists of a series of bytes, each byte is an operation, and each operation has a corresponding operation code. In smart contract code analysis, the source code in a smart contract is often converted to opcodes for better readability \cite{amani2018towards}\cite{chen2017under}.

\textbf{Ethereum transaction}: Ethereum transactions usually send messages from one account to another account with binary data or ETH (the native currency for the Ethereum platform) \cite{antonopoulos2014mastering}\cite{vujivcic2018blockchain}. 

\subsection{Smart Ponzi Scheme}
The control logic of a contract account is defined by its smart contract. Thus, criminals can implement the Ponzi scheme on Ethereum by including code related to the Ponzi scheme in smart contracts. Ponzi scheme contracts share a few patterns, and most new Ponzi scheme contracts are obtained by modifying existing ones. Specifically, the smart Ponzi scheme can be divided into four schemes, namely array-based pyramid scheme, tree-based pyramid scheme, handover scheme, and waterfall scheme \cite{Bartoletti2017Dissecting}. The array-based scheme refunds users based on their arrival order. Old users can make a profit when the funds raised from new users are sufficient. The tree-based scheme uses a tree structure to store the addresses of the Ethereum accounts, the root of the tree is the contract owner, and each user in the tree has a parent node. When a new user joins the program, his investment is used to refund his ancestors. The handover scheme only stores the address of the last user. If a new user wants to join the program, he needs to pay off the investment of the last user plus a fixed interest. The waterfall scheme refund users by dividing new investments among existing users. Starting with the first user, each user gets a fixed percentage of their investment in turn until the new investment is exhausted.

\subsection{Smart Ponzi Scheme Detection}
The detection of the Ethereum Ponzi scheme is a new topic emerging after the prosperity of the Blockchain. The existing smart Ponzi scheme detection methods can be divided into two categories, namely the code inspection-based method and the feature-based method. The code inspection-based detection method first converts the code in the smart contract into the corresponding opcode, and then analyze the control logic of the smart contract based on the operation code. Salvatore et al. manually examined 900 smart contracts on Ethereum to study the logic, life cycle, and financial impact of the smart contracts. They categorized the smart Ponzi scheme patterns into array-based pyramid schemes, tree-based pyramid schemes, handover schemes, and waterfall schemes. Nicola et al. \cite{atzei2017survey} provided a taxonomy of Ethereum smart contract security vulnerabilities based on where they were introduced (Solidity, EVM bytecode, or Blockchain) and listed a set of Blockchain attacks that can be implemented by smart contracts. Ting et al. \cite{chen2017under} defined several gas-costly smart contract programming patterns and developed a tool to automatically detect the patterns from the contract source code.

The feature-based smart Ponzi scheme detection method regards smart Ponzi scheme detection as a binary classification problem. A feature-based model uses a set of features to represent a contract account and then uses supervised learning to train a binary classifier to identify whether the account is a smart Ponzi scheme account based on the representation. The classifier can be random forests, neural networks, XGBoost, etc. The main difference between different feature-based approaches lies in how to define features. For instance, Massimo et al. \cite{Bartoletti2018Data} used 11 features to represent a Bitcoin address. These features include the time the account exists, the amount of Bitcoin received by the account, the ratio of incoming and outgoing transactions to/from the address, and so on. In \cite{Jung2019DataME}, the authors argued that opcodes in smart contracts can also be applied to contract account representations because they reflect the control logic of the account. The authors divided the account features into transaction-based features and code-based features. The transaction features are similar to the features used in \cite{Bartoletti2018Data}, and the code-based features of a contract are defined by the frequency of each opcode in the smart contract. The article also analyzed the importance of these features in the detection of smart Ponzi schemes.  Weili et al. \cite{Chenweili2018} examined the Ether flow graph and obtain three smart Ponzi scheme patterns, that is, a smart Ponzi scheme contract mainly paid to known accounts, lots of investment transactions had no refund, and some of the participants paid more than other investment accounts. According to the patterns, the authors devised a set of features for smart Ponzi scheme detection, such as the proportion of receivers who have invested before payment, the proportion of investors who received at least one payment. Compared with the features used in \cite{Bartoletti2018Data} and \cite{Jung2019DataME}, the features designed in \cite{Chenweili2018} can more accurately reflect the characteristics of the smart Ponzi scheme. Steven et al. \cite{FARRUGIA2020113318} designed 42 features to represent an account based on the analysis of transaction records and used XGboost as the classifier to detect smart Ponzi scheme. Fan et al. \cite{2021Al} imporved the performance of the feature-based smart Ponzi scheme detection method through eliminating imbalanced datasets.

%The authors also release the dataset used in the experiment.

Most of the feature-based approaches focus on designing new features for the representation of contract accounts. The problems with this method are: 1) The features used to represent the account are designed based on manual analysis of transaction data, which requires a lot of labor costs. The existing technology supports us to use artificial intelligence technology and massive data to save labor costs \cite{GeReserach2020}; 2) the risk of overfitting increases with the number of features; 3) feature-based methods are often poorly portable, and features designed for one environment may not suitable for another environment.

\subsection{Node Embedding}
\label{subsec_nodeEmbedding}
Node embedding is a technique for representing a network in a low-dimensional space by learning a continuous vector for each node in the network. The graph information is preserved in the vectors, and then graph algorithms can be computed efficiently on node vectors. In recent years, there have been a large number of methods proposed to learn efficient node embedding. One of the main differences between the various approaches is the information to be preserved. Some methods primarily preserve first-order proximity between nodes. These methods assume that if two nodes are connected, their node vector should be similar or close. For example, Locally Linear Embedding (LLE) \cite{roweis2000nonlinear}, Isomap \cite{balasubramanian2002isomap}, and DeepWalk \cite{perozzi2014deepwalk} are node embedding methods that preserve first-order proximity. LEE first constructs an affinity graph, where each node on the affinity graph is represented by its $K$ nearest neighbors, and then learns the low-dimensional vector of each node by solving the leading eigenvectors of the affinity matrix. The core idea of Isomap is similar to LEE. Isomap first obtains a distance matrix that records the distance between each pair of nodes of a given graph and then uses the eigendecomposition of the distance matrix as the embedding of the nodes in the graph. DeepWalk was inspired by word2vec \cite{mikolov2013efficient}, a model that generates word embedding. DeepWalk uses DFS to walk randomly on the network to capture the network structure, and each walk collects a list of connected nodes. The authors found that if each node is considered a word, then the node list is a sentence, and the embedding of a node can be calculated by word2vec. Other first-order proximity preserving network embedding methods include graph factorization\cite{goyal2018graph}, Eigenmap \cite{belkin2002laplacian}, Directed graph embedding \cite{chen2007directed}, etc.

The second-order proximity of a pair of nodes is defined by the similarity of their neighbors. There are some methods generate node embedding by preserving both first-order proximity and second-order proximity of network. These methods include Large-scale Information Network Embedding (LINE) \cite{tang2015line}, node2vec \cite{grover2016node2vec}, Structural Deep Network Embedding (SDNE) \cite{wang2016structural}, etc. LINE defines two different objective functions for the first and second-order similarity of the network. The node embedding generated by LINE can aware of its higher-order neighbors. This paper also proposed a sampling algorithm that enables LINE to be applied to large-scale networks. Node2vec uses both a depth-first search and breadth-first search to obtain the paths generated by random walks. These paths can capture not only the first-order proximity of the network but also the higher-order proximity of the network. Then, node2vec uses the same method as Deepwalk to generate node embeddings in the network. SDNE first uses a semi-supervised multi-layer model to capture the second-order proximity of the network, and then learns node embeddings by jointly optimizing the first-order proximity and the second-order proximity of the network.

In real life, graph nodes may be associated with external information such as text and labels. Some studies have tried to integrate the network structure and the external information of each node into the node embedding. For example, Cheng et al. \cite{yang2015network} first proved that DeepWalk is equivalent to matrix factorization. The authors then proposed text-associated deep walking (TADW) that integrates the text features of nodes into the network embedding through matrix decomposition. Xiaofei et al. \cite{sun2016general} proposed content-enhanced network embedding (CENE), which treats text information as a special kind of node to integrate text modeling and network structure modeling in a unified framework, and optimizes the probabilities of heterogeneous links. In \cite{tu2017cane}, the authors designed context-aware network embedding (CANE) that uses a mutual attention mechanism to generate context-aware embeddings for nodes. The node embedding generated by CANE not only contains the text and structure information of the nodes, but also the text and structure information of its neighboring nodes.

Many real-world networks, such as social network and biological networks, are dynamic and evolving. The node embedding of temporal graph has become the focal point of increasing research interests recently.  The existing temporal graph embedding methods can be divided into two categories, namely time-based methods and event-based methods. The time-based temporal node embedding method records the changes of the graph after each fixed time interval, and then computes the node embeddings in the graph based on these changes. For example, Lun et al. \cite{du2018dynamic} extended LINE to a dynamic setting. Given a sequence of network snapshots within a time interval, the model can generalize  to new vertex representation and update the most affected original vertex representations  during the evolvement of the network. Uriel et al. \cite{singer2019node} proposed a downstream task-aware temporal node embedding method. The algorithm uses static node embedding for initialization, then aligns the node representations at different timestamps, and eventually adapts to a given task in joint optimization. The event-based temporal node embedding method uses the events in the network to characterize the changing pattern of the network. For example, Yuan et al. \cite{zuo2018embedding} describe the evolution of a node by using temporal excitation effects exist between neighbors in the sequence.  They proposed a time node embedding method that integrates Hawkes process into network embedding to capture the influence of historical neighbors on current neighbors. In \cite{zhou2018dynamic}, the authors introduced a method for temporal node embedding based on triad, which is one of the basic units of the network. This method captures network dynamics and learns representations by modeling how a closed triad, which consists of three vertices connected, develops from an open triad that has two of three vertices not connected.

\section{Problem statement}
\label{sec_prob_state}
We consider smart Ponzi scheme detection in Ethereum with $|V|$ accounts. First, we model the Ethereum transaction network by constructing a temporal directed graph $G=(V, E, O, S) $ based on Ethereum transactions. The node $v \in V$ represents an account on Ethereum (it can be an externally owned account or a contract account). The edge $((v,u), w_{v,u}) \in E$ represents a transaction between the nodes $v$ and $u$, and the weight of the edge $w_{v,u}$ is the number of transactions from $v$ to $u$. The file $o \in O$ is the opcode of a smart contract. The interactive formation sequence $\{(v, u_1), (v, u_2),..., (v, u_N)\} \in S$ denotes the nodes have transactions with $u$ in turn. The interactive formation sequence describes how the transaction network changes over time.

Our task is to design a model which satisfies the following aspects.
\begin{enumerate}
\item The model takes $G = (V, E, O, S)$ as the input.
\item The model learns the evolution of nodes over time by the interactive accounts formation sequence.
\item The model outputs a boolean value $y$ which indicates whether $v$ is a smart Ponzi scheme account.
\item The prediction of $y$ should be as accurate as possible.
\end{enumerate}

\section{Methodology}
\label{sec_method}
\begin{figure*}[t]
    \centering
	\includegraphics[width=0.6\textwidth]{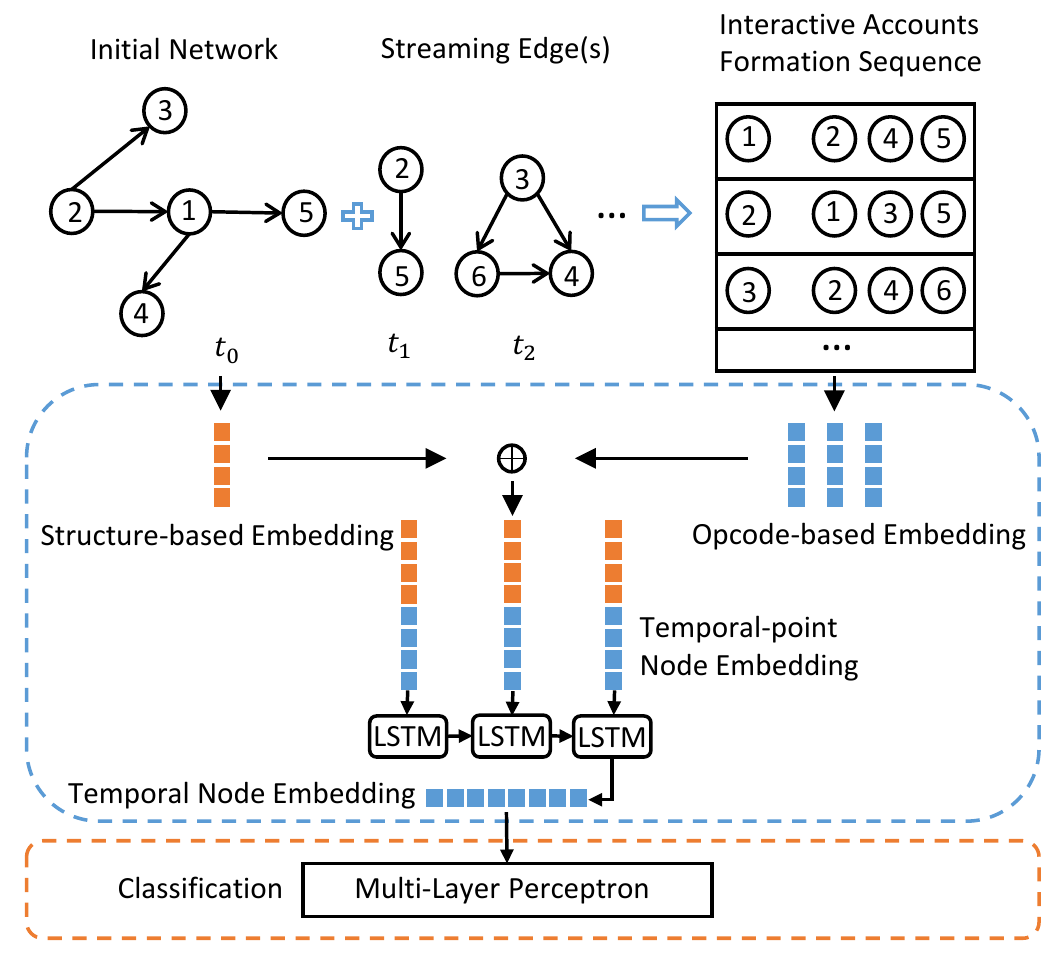}
	\caption{DSPSD Architecture}
	\label{fig_overview}
\end{figure*}

%The smart Ponzi scheme detection problem is formulated as a binary classification problem. To solve the problem, we first build a directed graph $G = (V, E, O, S)$ as that stated in section \ref{sec_prob_state}, and then we propose an algorithm that learns the evolution of nodes over time by the interactive accounts formation sequence and then delivers a binary classification that indicates whether the account of $v$ is a smart Ponzi scheme account.

As stated in section \ref{sec_prob_state}, the smart Ponzi scheme detection problem is formulated as a binary classification problem. In general, our idea is to use dynamic graph embedding to represent the structural information and temporal information of the transaction network and the opcode information of the account with a low-dimensional continuous vector, and then use a binary classifier to define whether the account is a smart Ponzi scheme account. The using of information from different source entitle the algorithm observe the account from different view and is beneficial for DSPSD generate a correct answer. 

As described in Section \ref{subsec_nodeEmbedding}, the existing temporal graph embedding methods can be divided into time-based methods and event-based methods. The data-driven smart Ponzi scheme detection system (DSPSD) we designed is an event-based approach. According to our observations, for a node in a transaction network, the related transactions often occur within a short period, and the network structure near this node does not change much in most of the time. Therefore, if time-based temporal node embedding is used, most of the network changes associated with the account occur in 1-2 timeslots, which is not sufficient to capture the evolution process of the transactions of the account.

\subsection{Overview}

The overall architecture of DSPSD is shown in Fig. \ref{fig_overview}. The detection process is divided into two steps, namely designing contract account embedding and classifying the embedding. In the first step, for a contract account, the transaction network structure, control logic, and dynamic information of the transaction network related to the account will be embedded into a continuous vector as a representation of the account. Specifically, assume a node $v$ has transactions with nodes $u_1$, $u_2$,..., $u_N$ in turn, then we call $\{v|u_1, u_2,..., u_N\}$ as the interactive account sequence of $v$. Let $t_i$ denote the time of the $i$-th event, we compute the representation of $v$ at $t_i$, denoted by $\vec v_{t_i}$, based on the transaction network $G_{t_i}$ at $t_i$. The vector $\vec v_{ti}$ is calculated from two aspects, that is, structure-based embedding and opcode-based embedding. After computing $\{ \vec v_{t1}, \vec v_{t2}, ..., \vec v_{tN} \}$, Long Short-Term Memory (LSTM) \cite{greff2017LSTM} is employed to learn the overall representation of $v$. Then, in the classification step, we put the learned representation of $v$, denoted by $\vec v$, into a multi-layer perceptron (MLP). The MLP outputs a Boolean value to indicate whether the account of $v$ is implementing a smart Ponzi scheme.

%

%the transaction network is $G_{t} = G_{t-1}+(v, u_t)$. We compute the temporal-point node embedding of $v$ at $t$ from two aspects: 1) the structure-based embedding which defines the representation of $v$ by the the topology of $G_t$; 2) the
%neighbour-aware opcode-based embedding which defines the representation of $v$ by the opcode of $v$ and the opcode of its neighbours. After computing the embedding of $u$ at each timestamp, we employ Long Short-Term Memory (LSTM) \cite{DBLP:journals/tnn/GreffSKSS17} to learn the overall embedding of $u$. In the classification step, we use a Multi-Layer Perceptron (MLP) \cite{DBLP:journals/tnn/RuckRKOS90} as our classifier. The MLP takes the embedding of $v$ as input, and outputs an boolean value to indicates whether $u$ is an account implementing smart Ponzi scheme.

%Accordingly, we divide DSPSD into two modules, a temporal representation module and a classification module. The temporal representation module consists of two components, namely temporal-point node embedding and node embedding via the interactive accounts formation sequence, which are introduced in Section \ref{temporal-point_emb} and Section \ref{temporal_emb}, respectively. The classification module is introduced in Section \ref{sec_classification}.

\subsection{Temporal-point Node Embedding}
\label{temporal-point_emb}
\begin{figure*}[t]
        \centering
        \includegraphics[width=\textwidth]{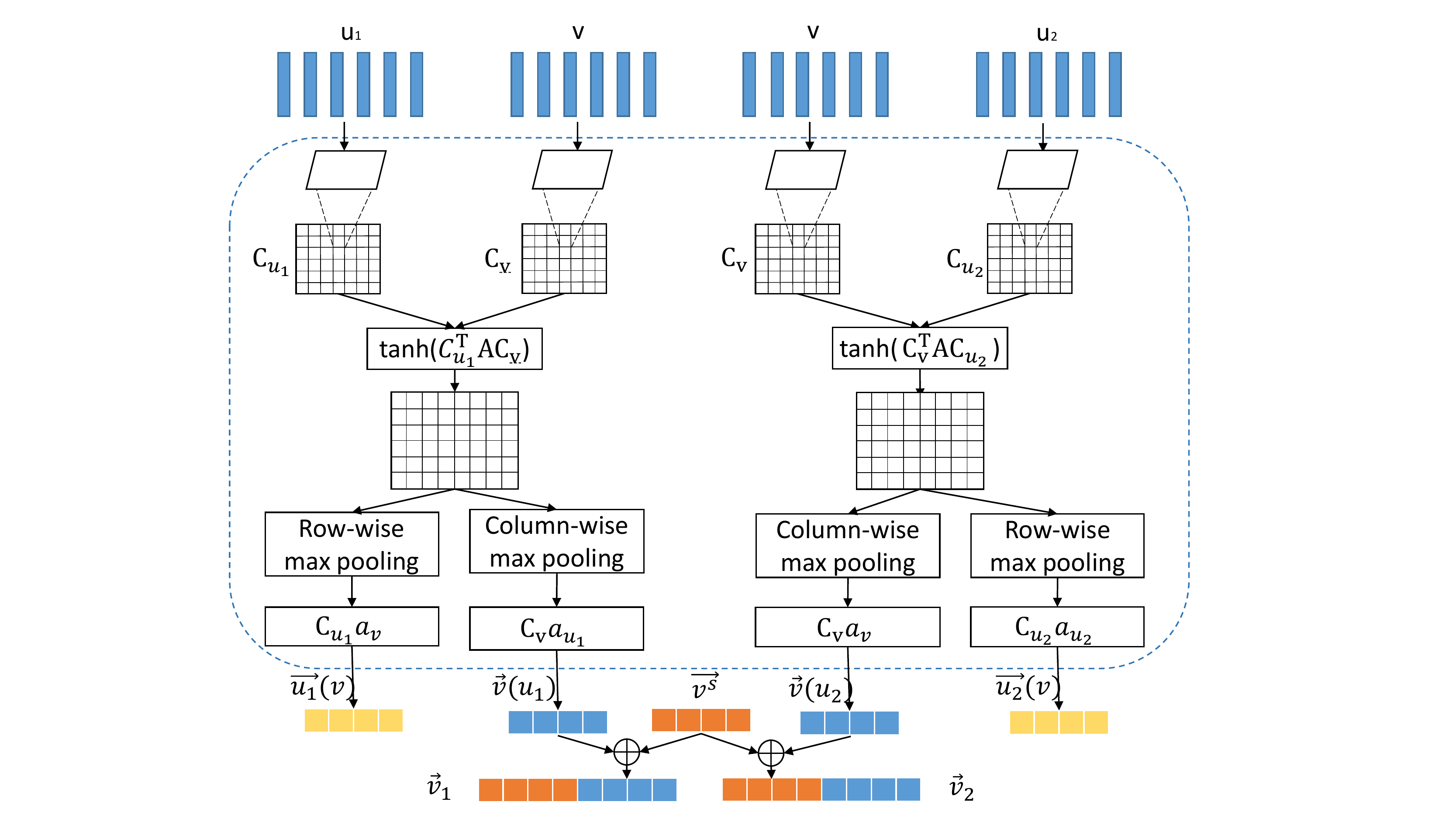}
        \caption{Process of Computing Opcode-based Embedding}
        \label{att_pooling}
\end{figure*}
The temporal-point node embedding is to compute the embedding of a node at a specific timestamp. Let $G_{t_i} = (V_{t_i}, E_{t_i})$ denote the transaction network at $t_i$, we compute the temporal-point node embedding of $v$ from two aspects. First, the structure-based embedding that defined by the topology of $G_{t_i}$. Second, the interactive account-aware opcode-based embedding that defined by the control logic of $v$ and its interactive accounts.

%the structure-based embedding which defines the representation of $v$ by the the topology of the transation network at $t$; 2) the neighbourhood-aware opcode-based embedding which defines the representation of $v$ by the opcode of $v$ and the opcode of its neighbours at $t$.

\label{sec_str-emb}
\textbf{Structure-based Embedding}: The structure-based embedding is to generate a vector $\vec {v}_{t_i}^s \in R^d_s$ according to the topology of $G_{t_i}$, where $d_s$ is the dimension of the embedding. The core idea is to embed the information of the edge weights into the node representations. Concretely, for an edge $((v_{t_i}, x), w_{v_{t_i},x}) \in E_{t_i}$, the conditional probability of $x$ generated by $v_{t_i}$ is defined as
\begin{equation}
\label{eq_joint}
    p(x|v_{t_i}) = \frac{exp(\vec {x^s}^T \cdot \vec v_{t_i}^s)}{\sum_{z \in V_{t_i}}exp({\vec z}^T \cdot \vec v_{t_i}^s)}
\end{equation}
where $\vec x^s$, $\vec v_{t_i}^s$ are the structure-based vectors of $x$ and $v_{t_i}$, the symbol $T$ represents matrix transposition.

Then, for edge $e = (v_{t_i}, x)$, the objective is to minimize
\begin{equation}
\label{eq_structure_loss_ti}
	\mathcal{L}_{t_i}^s(e) = - w_{v_{t_i},x}\log p(x|v_{t_i}).
\end{equation}
where $w_{v_{t_i}, x}$ represents the weight of $(v_{t_i}, x)$.

\label{sec_op-emb}
\textbf{Opcode-based Embedding}: The opcode-based embedding of $v_{t_i}$, denoted by $\vec v_{t_i}^o$, is generated from three aspects. First, the control logic of $v_{t_i}$. Second, the control logic of the interactive accounts of $v_{t_i}$ in $G_{t_i}$. The interactive accounts of $v_{t_i}$ refers to the accounts that have transactions with $v_{t_i}$ before $t_i$. Third, the formation sequence of the interactive accounts of $v_{t_i}$.

%Therefore, we use a interactive account-aware embedding based on the associated opcode information. Each node can have its own points of focus about a specific node. For

%Specifically, we assume that one account node usually shows different aspects when interacting with different neighbor accounts, and should own different embeddings respectively. TFor an edge $(v_t, u_t) \in E_t$, the opcodes of $u_t$ will affect $\vec v_t^o$ by a two-way attention mechanism.

Assume $U = \{ v_{t_i}|u_1, u_2,..., u_k \}$ are the interactive accounts of $v_{t_i}$ before $t_i$, the computation of opcode-based embedding of $v_{t_i}$ , denoted by $\vec {v}_{t_i}^o$ can be divided into 3 steps: 
\begin{enumerate}
    \item Computing a representation for the control logic of $v_{t_i}$ and the nodes in $U$;
    \item Computing the interactive account-aware opcode-based representation of $v_{t_i}$ based on each node in $U$;
    \item Integrating the representations obtained from step 2.
\end{enumerate}

The first step is to compute the control logic representation of each node. The control logic of the node is determined by the operation code of the contract account. Therefore, the computation of the control logic representation can be converted into a text encoding problem, that is, encoding the content of the opcode (text information) into a continuous vector. In DSPSD, we use a CNN-based encoding method, which is widely used in text coding\cite{zhang-wallace-2017-sensitivity}\cite{jacovi-etal-2018-understanding}, to obtain the representation. In the encoding of the text, each word in the text is usually represented by a word vector, and then the word vectors of all the words in the text are stacked into a matrix. CNN captures the features between multiple consecutive words through the convolution kernel and shares the weights when calculating the same type of features, thereby capturing the local semantic dependence between words. Specifically, the representation of the control logic of node $v$ is computed as follows. First, prepare an opcode dictionary. For any opcode, it has the corresponding random generated vector representation $\vec o' \in R^{d'}$ in the opcode dictionary. Assuming that the length of the contract of $v$ is $L$, the control logic matrix of node $v$, denoted by $X_v \in \mathbb{R}^{L \times d'}$, can be produced by looking up the opcode dictionary and stacking the vectors. For an EOA without associated contracts, its $X_v$ is a matrix of all zeros. Then, we employ a convolutional layer to extract the local features of $X_v$. In the convolutional layer, a learnable filter $F \in \mathbb R^{r \times d'}$, $r$ is a user-defined parameter, are applied over $X_v$ to capture the local multi-gram information. For instance, a new feature $p_i$ is extracted from $X_v[i:i+r-1]$ according to the following formula:
\begin{equation}
\label{eq_cnn_filter}
	p_i= f(F \cdot X_v[i:i+r-1]+b)
\end{equation}
where $\cdot$ is element-wise multiplication, $b \in \mathbb R$ is a bias term and $f$ is a non-linear function such as the hyperbolic tangent. The filter will be applied to the representations of whole node matrix $X_v$ via a sliding window to establish the feature map, i.e., the out of the convolutional layer $\vec{q} =[p_1,p_2,...,p_{L-r+1}]$. Let $l=L-r+1$, $\vec q \in \mathbb{R}^l$.

For $d_o$ filters with the same length, the generated feature maps can be rearranged as feature representations of node $v$ for each window:
\begin{equation}
	 C_v = [\vec {q}_1;\vec {q}_2;...;\vec {q}_{d_o}]
\end{equation}
Here, semicolons represent column vector concatenation and $\vec {q_i}$ is the feature map generated with the $i$-th filter, $C_v \in R^{d_o \times l}$.

In this way, we can compute the representation of the control logic of $v_{t_i}$ and the nodes in $U$.

The second step is to compute the interactive account-aware opcode-based representation of $v_{t_i}$. We assume that $v_{t_i}$ has different aspects when interacting with different nodes in $U$. To achieve this, we employ mutual attention \cite{2016arXiv160203609D} to obtain interactive account-aware embedding. The mutual attention enables the pooling layer in CNN to be aware of the vertex pair in an edge. Specifically, to generate the representation of $v_{t_i}$ based on $u_j$, we compute a correlation matrix $D \in \mathbb{R}^{l \times l}$ by using an attentive matrix $A \in \mathbb{R}^{d_o \times d_o}$ as follows:
\begin{equation}
	D = tanh(C_{v_{ti}}^{T}AC_{uj})
\end{equation}
where $C_{v_{t_i}}$ and $C_{u_j}$ are the representation of the control logic of $v_{t_i}$ and $u_j$, and $D[m,n]$ in $D$ represents the pair-wise correlation score between $C_{v_{t_i}}[m]$ and $C_{u_j}[n]$.

Afterward, we conduct a max-pooling operation along rows and columns of $D$ to generate opcode-based attention vectors for node $v_{t_i}$ and node $u_j$ separately.

The attention vectors are denoted by ${\vec a}_{v_{ti}}$ and $\vec {a}_{u_j}$, where $\vec {a}_{v_{ti}}, {\vec a}_{uj} \in \mathbb{R}^d_o$. The interactive account-aware node representation of $v_{ti}$ based on $u_j$ can be computed as:
\begin{equation}
\label{v_emb}
	\vec v_{t_i}(u_j)=C_{v_{t_i}}\vec a_{u_j}
\end{equation}
The interactive account-aware node representation of $u_j$ based on $v_{t_i}$ is:
\begin{equation}
\label{u_emb}
	\vec u_j(v_{t_i})=C_{u_j}\vec a_{v_{t_i}}
\end{equation}
For $e = (v_{t_i}, u_j)$, the objective is to minimize 
\begin{equation}
	\mathcal{L}_{t_i}^o(e)= - w_{v_{t_i},u_j}\log p(u_j(v_{t_i})|v_{t_i}(u_j))
\end{equation}
where
\begin{equation}
p(u_j(v_{t_i})|v_{t_i}(u_j)) = \frac{exp(\vec u_j(v_{t_i}) \cdot \vec v_{t_i}(u_j))}{\sum_{z \in V_{t_i}}{exp(\vec v_{t_i}(z) \cdot \vec z(v_{t_i}))}}
\end{equation}

The third step is to integrate $\{ \vec v_{t_i}(u_1), \vec v_{t_i}(u_2),..., \vec v_{t_i}(u_j) \}$. We obtain $\vec v_{t_i}^o \in \mathbb{R}^{k \times d_o}$ by concatenating the vectors:
\begin{equation}
\label{eq_opcode_ti}
    \vec v_{t_i}^o = [v_{t_i}(u_1): v_{t_i}(u_2), ..., : v_{t_i}(u_j)]
\end{equation}

The overall loss of DSPSD at time $t_i$ is
\begin{equation}
\label{eq_overall_loss_ti}
    \mathcal{L}_{t_i} =  \sum_{e \in E_{t_i}} \mathcal{L}_{t_i}^s(e) + \mathcal{L}_{t_i}^o(e)
\end{equation}

One of the problems is that the loss function contains several conditional probabilities that are computationally expensive in optimization. We can employ negative sampling \cite{mikolov2013distributed} to solve the problem. Afterward, optimization methods such as stochastic gradient descent \cite{rumelhart1988learning}, Adam \cite{kingma2014adam}, can be utilized to minimize $\mathcal{L}_{t_i}$ and train the system.

%and transform Eq. \ref{overall_loss} into the following form:

%Thus, according to the degree distribution $P_n(v) ∝ {d_v}^{3/4}$, where $d_v$ is the degree for node $v$, we employ negative sampling \cite{DBLP:conf/nips/MikolovSCCD13} and transform Eq. \ref{overall_loss} into the following form:
%\begin{equation}
%\label{eq_neg_loss}
%    \log{\sigma (\vec u_1^{T} \cdot \vec v)} + \sum_{i=1}^m \mathbb E_{v_i\sim P_{(v)}}[\log{\sigma (\vec u_1^{T} \cdot \vec z)}]
%\end{equation}
%where $m$ is the number of negative samples and $\sigma$ represents the sigmoid function. We minimize the loss in Eq. \ref{eq_neg_loss} by stochastic gradient descent.

In order to better explain the algorithm, we illustrate the process of computing opcode-based embedding of a node $v$ with interactive account formation sequence $\{v|u_1, u_2\}$ in Fig. \ref{att_pooling}. First, the control logic representation of $v$, $u_1$, and $u_2$ are calculated based on the opcodes of the nodes. Then, the interactive account-aware opcode-based embedding of $v$ is computed according to $u_1$ and $u_2$. Finally, the overall opcode-based embedding of $v$ is obtained by concatenating  the interactive account-aware opcode-based embedding of $v$ based on $u_1$ and $u_2$.

\subsection{Node Embedding over Interactive Accounts Formation Sequence}
\label{temporal_emb}

For a node $v$ with interactive account formation sequence $\{v | u_1, u_2,..., u_N\}$, we can compute its temporal-point embedding of each event $\{\vec v_{t_1}, \vec v_{t_2}, ..., \vec v_{t_N}\}$, and the order in the sequence reflects the way that $v$ evolves over time. To obtain an embedding with interactive accounts formation sequence information, we need to aggregate node representations at different timestamps, that is, merging $\{\vec v_{t_1}, \vec v_{t_2}, ..., \vec v_{t_N}\}$ into a vector $\vec v$. This task can be transformed into a sequence modeling problem. The state of the last unit of the sequence model can be regarded as a comprehensive representation of the input at different timestamps. We use Long Short-Term Memory (LSTM) to achieve this aggregation. LSTM is a widely used and effective sequence modeling method\cite{2015Modelling}\cite{2019A}. It improves the long-term dependency problem in Recurrent Neural Networks (RNN), and the performance of LSTM is usually better than RNN and Hidden Markov Model (HMM). With the increase of data, LSTM uses a self-cycling structure with gates to transfer the last state to the current input.  The function of the gate is to decide what information to forget or what information to continue to transmit. Concretely, we the embeddings $\{\vec v_{t_1}, \vec v_{t_2}, ..., \vec v_{t_N}\}$ one by one into an LSTM, and the last memory unit represents the overall embedding of $v$, i.e., $\vec v = \vec h_N$ ($h_N \in R^{d_s + d_o}$). The content of the $i$-th memory unit $\vec h_{t_i}$ is defined by the $i$-th temporal-point embedding $\vec v_{t_i}$ and the output of last memory unit $\vec h_{t_{i-1}}$. There are three gates in the unit, namely the input gate $in_{t_i}$, the forget gate $f_{t_i}$ and the output gate $o_{t_i}$. The formulas for the three gates are as follows.

%For a node $v$ with interactive account formation sequence $\{v | u_1, u_2,..., u_N\}$, we can compute its temporal-point embedding of each event $\{\vec v_{t_1}, \vec v_{t_2}, ..., \vec v_{t_N}\}$. Then, we can create a representation for each node based on its historical embeddings. Concretely, we use Long Short-Term Memory (LSTM) to reduce sequence data into a vector. LSTM is a recurrent neural network used to capture long term order dependencies between elements in sequences. For $\{\vec v_{t_1}, \vec v_{t_2}, ..., \vec v_{t_N}\}$, the order in the sequence reflects the way that $v$ evolves over time. Therefore, we input the embeddings one by one into an LSTM, and the last memory unit represents the overall embedding of $v$, i.e., $\vec v = \vec h_N$ ($h_N \in R^{d_s + d_o}$). Specifically, the content of the $i$-th memory unit $\vec h_{t_i}$ is defined by the $i$-th temporal-point embedding $\vec v_{t_i}$ and the output of last memory unit $\vec h_{t_{i-1}}$. There are three gates in the unit, namely the input gate $in_{t_i}$, the forget gate $f_{t_i}$ and the output gate $o_{t_i}$. The formulas for the three gates are as follows.
\begin{equation}
\label{eq_lstm_start}
    in_{t_i}=\sigma(W_i[\vec h_{t_{i-1}}, \vec v_{t_i}]+\vec b_i)
\end{equation}
\begin{equation}
    f_{t_i}=\sigma(W_f[\vec h_{t_{i-1}}, \vec v_{t_i}]+\vec b_f)
\end{equation}
\begin{equation}
    o_{t_i}=\sigma(W_o[\vec h_{t_{i-1}}, \vec v_{t_i}]+\vec b_o)
\end{equation}
where $W_x$, $\vec b_x$ are weights and biases for respective gates.
Then, the content of $h_{t_i}$ is defined by:
\begin{equation}
    \widetilde S_{t_i}=tanh(W_S[\vec h_{t_{i-1}},\vec v_{t_i}]+\vec b_S)
\end{equation}
\begin{equation}
    S_{t_i}=f_{t_i} \times S_{t_{i-1}}+in_{t_i} \times \widetilde S_{t_{i-1}}
\end{equation}
\begin{equation}
\label{eq_lstm_end}
    h_{t_i}=o_{t_i} \times tanh(S_{t_i})
\end{equation}
where $S_{t_i}$ is the unit state can be updated.

In reality, the number of interactive accounts for each node is different. However, We have to make them of equal length before we take them as input for Long Short-Term Memory (LSTM). Rather than simply truncating, we employ unweighted averaging for nearby temporal-point embedding to reduce the sequence length while keeping the whole interaction process. For some short sequence, we simply pad them with zeroes in the ending. After padding, each temporal-point node embedding is fed as input to an LSTM memory cell. The last memory cell of the LSTM represents the final temporal embedding of the node, optimizing for the classification.

%In our work, different source node $v$ have different interactive accounts formation sequences with different lengths. Especially, some Ponzi scheme accounts interact with neighbor nodes more frequently than ordinary contract accounts such that the length is longer. 

\subsection{Classification}
\label{sec_classification}
\begin{figure}[t]
        \centering
        \includegraphics[width=0.45\textwidth]{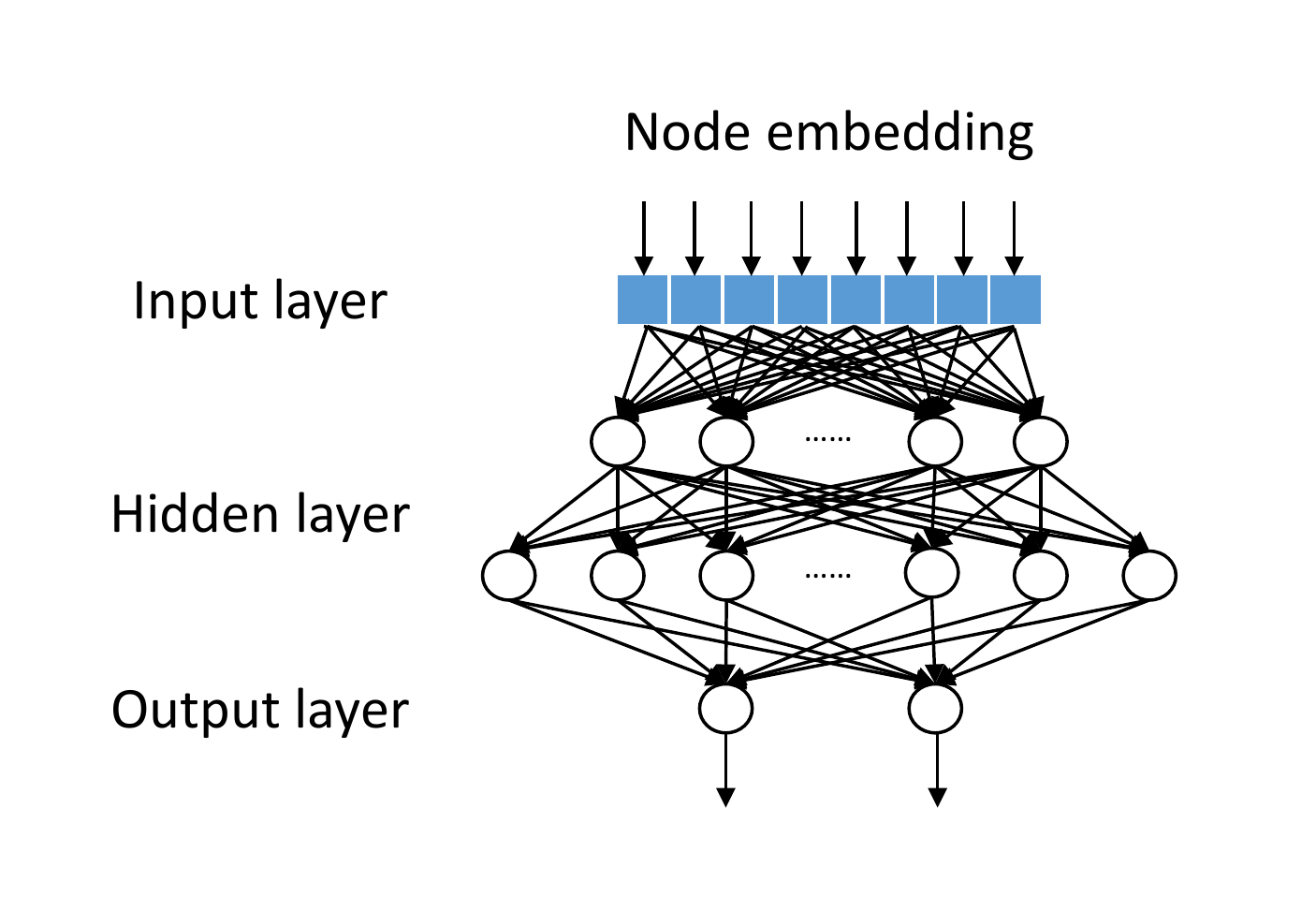}
        \caption{A Multi-Layer Perception}\label{fig_MLP}
\end{figure}

The embedding of a node will be input to a binary classifier, and the output of the classifier indicates whether the node is a smart Ponzi scheme account. In DSPSD, we leverage a Multi-Layer Perceptron (MLP) as the classifier. MLP is a simple but effective neural network structure, which is widely used in classification problems.

Figure \ref{fig_MLP} presents an MLP that has two hidden layers and an output layer. An MLP can be divided into the input layer, hidden layer, and output layer. Each layer uses a sigmod function as the transfer function. Let $f_i$, $f_h$, and $f_o$ denotes the transfer function of the input layer, the hidden layer, and the output layer, the MLP works as follows.

The input layer of the MLP take $\vec v$ as its input.
\begin{equation}
\label{eq_mlp_start}
	\vec y_{in}=f_{in}(W_{in} \vec v+\vec b_{in})
\end{equation}
The output of the input layer $\vec y_{in}$ will be the input of the hidden layer.
\begin{equation}
	\vec y_h=f_h(W_h\vec y_{in}+\vec b_h)
\end{equation}
Then, the output layer is
\begin{equation}
	\vec y_o=f_o(W_{out}\vec y_h+\vec b_{out})
\end{equation}
Finally, we use a softmax function to obtain the classification result.
\begin{equation}
	\hat y = softmax(W_m\vec y_o+\vec b_m)
\end{equation}
where $W$ is a learnable weight matrix and $\vec b_m \in \mathbb{R}^{d_s + d_o}$ is a bias vector. Then we can get a boolean value indicates whether the input node is a smart Ponzi scheme account according to $\hat y$.

Assume the nodes in the network is $\{v_1, v_2,..., v_{|V|}\}$, $\hat y_i$ is the softmax result of $v_i$, $y_i$ is the golden label of $v_i$, the loss of the MLP is in Eq. \ref{eq_class_loss}.
\begin{equation}
\label{eq_class_loss}
\begin{split}
	\mathcal{L}_c=\sum_{i=1}^{|V|} [y_i \ln{(1+e^{-(\hat{y_i})})}+(1-y_i)\\
	\ln{(1+e^{(\hat{y_i})})}]
	+ \lambda {\left \| \theta \right \|}_2^2
\end{split}
\end{equation}
where $\theta$ contains all the parameters of the network and $\lambda {\left \| \theta \right \|}_2^2$ is the regulation term. The objective function $\mathcal{L}_c$ in Eq. \ref{eq_class_loss} measures how predictive the model is on the training data and the regulation term helps to avoid over-fitting.

To recap, the overall process of DSPSD is as follows:
\begin{enumerate}[Step 1:] 
\item Construct a transaction network $G=(V, E, O, S) $ according to the transaction data.
\item For each node $v \in V$ and its corresponding account formation sequence $s$ ($s = \{v|u_1,...,u_N\}$), construct $G_v = \{G_{t_1}...G_{t_N}\}, G_{t_i} = (V_{t_i}, E_{t_i})$;
\item Computes the structure-based embedding of $v_{t_i}$ according to Eq. \ref{eq_joint} and Eq. \ref{eq_structure_loss_ti};
\item Computes the opcode-based embedding of  $v_{t_i}$ according to Eq. \ref{eq_cnn_filter}-\ref{eq_opcode_ti};
\item Uses Eq. \ref{eq_overall_loss_ti} as the loss function to compute the representation of $v$ at $t_i$;
\item Computes $\vec v$ by aggregating $\{\vec v_{t_1},..., \vec v_{t_N}\}$ through Eq. \ref{eq_lstm_start}-\ref{eq_lstm_end}.
\item Classify $\vec v$ according to Eq. \ref{eq_mlp_start}-\ref{eq_class_loss} to determine whether the corresponding account has implemented a smart Ponzi scheme.
\end{enumerate}
\section{EXPERIMENT}
We used Ethereum's real transaction data to evaluate the effectiveness and efficiency of DSPSD. The data used in this paper is the same as that used in \cite{Chenweili2018}\footnote{The data is available at ibase.site/scamedb}. The data is part of the data collected from http://etherscan.io before May 7, 2017. Each of the records contains the payment account ID, recipient account ID, transaction time, transaction value, etc. There are 1251 normal contract accounts and 131 smart Ponzi scheme accounts. 

%There are a total of 692,518 transactions involving 35,996 EOAs, 438 normal contract accounts, and 50 smart Ponzi scheme accounts. 

Our model training has two parts. The first part is to generate node embeddings based on the transaction network structure and contract opcodes. In this part, we use all the transaction data to construct the transaction network, and use the edges in the transaction network as supervisory signals to train the model. The second part is classification training. In this part, we only focus on contract accounts and use the labeled data of contract accounts as supervisory signals.  We used 10-fold cross-validation to obtain the results. That is, we divided the dataset into 10 parts, of which 9 parts were used for training and the remaining part was used for testing. This process can be repeated 10 times, with different test data each time, and the final result is the average result of 10 experiments.

%and compute node embeddings according to Eq. \ref{overall_loss}
%based on Eq. \ref{eq_class_loss}
%First, we randomly select 80\% of the labeled data of the contract account as training data and the remaining 20\% as test data. Then, we train our model on the training data for 200 epochs, and select the parameters of the round with the smallest loss as the parameters of our model. Finally, we run the trained model on the test dataset to obtain the test results. 

%We build a transaction network based on these transactions. In the process of training MLP, we splits 80\% of the contract account data selected randomly into training set and the remaining 20\% sample into test data set.

%We randomly select the transactions from 6894 EOAs, 1001 normal accounts and 105 smart Ponzi scheme accounts as our training set, and the rest of the data as the test set. Overall, we have XXX transactions of 8000 accounts for training and XXXX transactions from 2000 accounts for test. More information about the dataset is listed in Table \ref{dataset} \footnote{We will publish the source code of the experiment after this article is accepted}. 

In the experiment, we first compared the performance of DSPSD with several baseline methods. For all baseline methods, we first generate a representation for each contract account through embedding or feature extraction and then input the representation into a classifier to detect whether the corresponding account is a smart Ponzi scheme account. We evaluated the performances from two aspects. First, we compared the effects of using different methods to generate contract account representations. Second, we compared the detection results of different classifiers. In addition, we visualized different types of embeddings and opcodes for contract accounts.

\section{Parameters Setting}
We initialized all the embeddings by randomly sampling from a uniform distribution in [-0.1, 0.1]. The vector length $d_o = d_s = d' = 100$. The width of CNN filters $r$ was set to 2, and the number of convolutional feature maps and the attentive matrix size $l$ were set to 100. An LSTM layer of size 32 follows with a dropout value of 0.75 and using hyperbolic tangent as the activation function. The MLP used in our experiment consists of 3 layers, each layer contains 32 or 64 hidden units. The model parameters are regularized with L2 regularization. We used stochastic gradient descent as the optimizer, and the range of the learning rate is set to [0.001, 0.01]. The maximum length of opcodes in a smart contract is set to 300. We train all the models in batches with a size of 64.

%In our experiment, we initialized all the embeddings by randomly sampling from uniform distribution in [-0.1, 0.1]. The embedding vector length $d$ and the initialization embedding length $d'$ both were set to 100. The width of CNN fiters $r$ was set to 2, and the number of convolutional feature maps and the attentive matrix size $l$ were set to 100. An LSTM layer of size 32 follows with a dropout value of 0.75 and using hyperbolic tangent as activation function. The MLP used in our experiment consists of 3 layers, and the units in hidden layer is set to 32 or 64. The model parameters are regularized with an L2 regularization. We used stochastic gradient descent as the optimizer, and the range of the learning rate is set to [0.001, 0.01]. The maximum length of opcodes in a smart contract is set to 300. We train all the models in batches with size of 64.

\subsection{Evaluation Metrics and Baseline Method}

Our experiment compares the impact of different methods on the detection results from two aspects, namely embedding method and classification method. The results were evaluated by precision, recall, and F-score. Precision $P$ is used to measure the exactness or quality of the method, and $P$ is defined as the number of correctly predicted positive items divided by the total number of predicted items. The recall $R$ is a measure of completeness or quantity, and $R$ is defined as the number of correctly predicted positive items divided by the total number of positive items in the dataset. F-score is a measure of the overall performance of a model, $F=\frac{2 \times P \times R}{P+R}$.

The following method serves as our baselines.
\begin{itemize}
\item \textbf{LINE + SVM}: LINE\cite{tang2015line} is a node embedding algorithm that preserves the first-order proximity and the second-order proximity of a network. The vector generated by LINE contains only the structural information of the static network. We use LINE to compute the embedding of each contract account and then use Support Vector Machine (SVM) \cite{nicholas2009time} to identify smart Ponzi scheme accounts. The support vector machine classifier performs classification by achieving the hyperplane that enlarges the border between two categories.

\item \textbf{LINE + MLP}: We use LINE to compute the embedding of each contract account, and then use MLP to identify smart Ponzi scheme accounts. MLP is a basic type of feedforward neural network that consists of an input layer, several hidden layers, and an output layer (Figure \ref{fig_MLP}).

\item \textbf{Control logic representation + SVM}: In order to generate the control logic representation of the node, we do not consider the structural information in the training, that is, remove $L_{t_i}^s(e)$ from Eq. \ref{eq_overall_loss_ti}. The control logic representation can then be input into a trained SVM to identify if the account is a smart Ponzi scheme account.

%As described in Section \ref{sec_method}, the control logic matrix of a node is a matrix defined by the opcodes of the node. This matrix is input to the CNN layer to extract features, and the output is flattened into a vector as the control logic representation of the node. The control logic representation can then be input into a trained SVM to identify if the account is a smart Ponzi scheme account.

%CNN\cite{} is a kind of neural networks for text modeling, which can capture the local logic dependency among opcodes. Taking the opcodes of an smart contract account as input to get opcode-only static node embedding, we use SVM to identify smart Ponzi scheme accounts.

\item \textbf{Control logic representation + MLP}: We input the control logic representation of a node into a trained MLP to identify if the node is representing a smart Ponzi scheme account.

%We use CNN to calculate the embedding of each contract account. Then the embeddings were input into a MLP to get the classification result.

\item \textbf{CANE + SVM}: For each node, CANE\cite{tu2017cane} integrates the network structure, text information associated with the node into a continuous vector. CANE has separate objective functions for structure-based embedding and text-based embedding. In our experiment, the opcodes of a node serve as the text information of the node. Then, we use SVM to classify the vectors generated by CANE. 

\item \textbf{CANE + MLP}: We use CANE to compute the embeddings of the contract accounts and use MLP to classify the embeddings.

\item \textbf{DPSE}: DPSE is a smart Ponzi scheme detection methods described in \cite{Chenweili2018}. DPSE uses account features and code features to represent contract accounts, and then uses Random forest (RF) as a classifier to identify whether the account is a smart Ponzi scheme account. Random forest is a classic machine learning tool based on aggregating the output of a collection of decision trees.
%When only transaction network is available, DPSE only can used account features, which are structure-based.
\end{itemize}

\begin{table*}[ht]
\centering
\caption{Smart Ponzi scheme Detection Result}
\renewcommand\arraystretch{1.5}
\begin{tabular}{|l|l|c|c|c|}
\hline
Algorithm & Embedding Description & Precision & Recall & F-score \\
\hline
LINE+SVM & structure-only, static node embedding & 0.67 & 0.32 & 0.43 \\
LINE+MLP & structure-only, static node embedding & 0.91 & 0.4 & 0.56 \\
Control logic representation+SVM & opcode-only, static node embedding & 0.92 & 0.61 & 0.73 \\
Control logic representation+MLP & opcode-only, static node embedding & 0.87 & 0.75 & 0.81 \\
CANE+SVM & structure + opcode, static node embedding & 0.87 & 0.83 & 0.85 \\ 
CANE+MLP & structure + opcode, static node embedding & 0.92 & 0.83 & \underline{0.87} \\
DPSE (structure-only)   & structure-only, feature-based & 0.74 & 0.32 & 0.44 \\
DPSE (structure + opcode) & structure + opcode, feature-based & \underline{0.94} & 0.81 & 0.86 \\
\hline
Our Embedding + SVM & structure + opcode, temporal node embedding & 0.90 & \underline{0.84} & \underline{0.87} \\
DSPSD (Our Embedding + MLP) & structure + opcode, temporal node embedding & \textbf{0.98} & \textbf{0.85} & \textbf{0.91} \\
\hline
\end{tabular}\label{result}
\end{table*}

% \begin{table*}[t]
% \centering
% \caption{Smart Ponzi scheme Detection Result}
% \renewcommand\arraystretch{1.5}
% \begin{tabular}{|l|l|c|c|c|}
% \hline
% Algorithm & Embedding Description & Precision & Recall & F-score \\
% \hline
% LINE + SVM & structure-only, static node embedding & 0.8 & 0.4 & 0.53 \\ 
% LINE + MLP & structure-only, static node embedding & 0.83 & 0.5 & 0.63 \\ 
% Control logic representation + SVM & opcode-only, static node embedding & 0.8 & 0.8 & 0.8 \\
% Control logic representation + MLP & opcode-only, static node embedding & 0.88 & 0.8 & 0.84 \\
% CANE + SVM & structure + opcode, static node embedding & 0.93 & 0.8 & 0.86 \\ 
% CANE+MLP & structure + opcode, static node embedding & 0.94 & 0.88 & 0.91 \\ 
% DPSE (structure-only)   & structure-only, feature-based & 0.64 & 0.2 & 0.3 \\
% DPSE (structure + opcode) & structure + opcode, feature-based & 0.95 & 0.69 & 0.79 \\
% \hline
% Our Embedding + SVM & structure + opcode, temporal node embedding & 0.91 & \textbf{1} & 0.95 \\
% DSPSD (Our Embedding + MLP) & structure + opcode, temporal node embedding & \textbf{0.98} & \textbf{1} & \textbf{0.99} \\
% \hline
% \end{tabular}\label{result}
% \end{table*}

\subsection{Smart Ponzi Scheme Detection Result}
\begin{figure*}[h]
    \centering
        \subfigure{
            \begin{minipage}[t]{0.3\linewidth}
            \centering
            \includegraphics[width=3.5in]{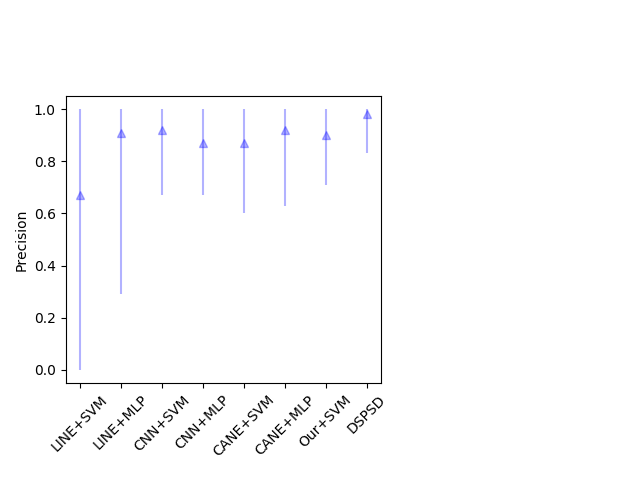}
            \end{minipage}%
        }%
        \subfigure{
            \begin{minipage}[t]{0.3\linewidth}
            \centering
            \includegraphics[width=3.5in]{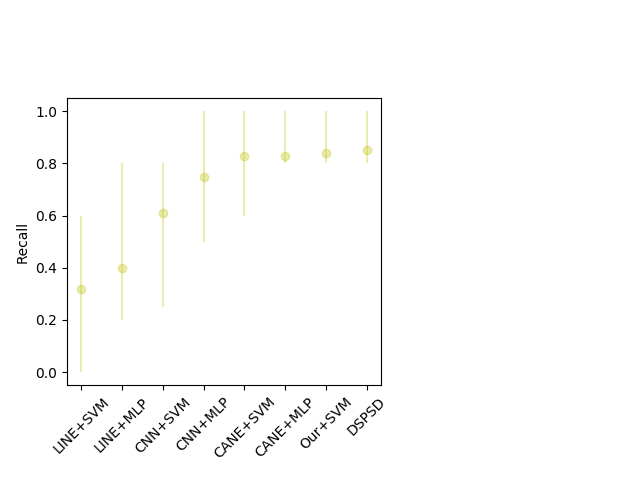}
            \end{minipage}
        }%
         \subfigure{
            \begin{minipage}[t]{0.3\linewidth}
            \centering
            \includegraphics[width=3.5in]{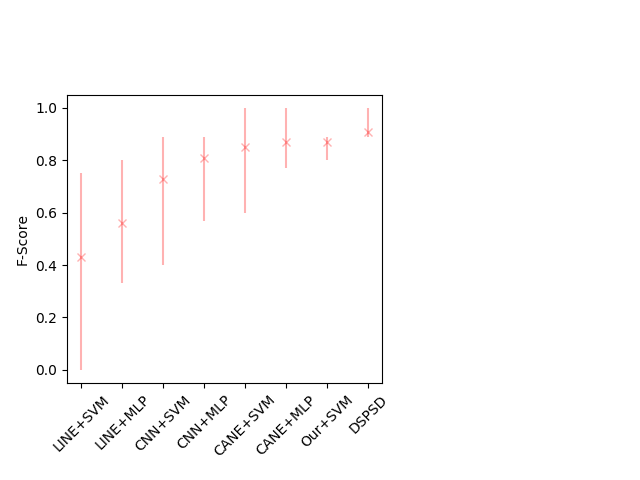}
            \end{minipage}%
        }%
    \centering
    \caption{Result Range of 10-fold Cross Validation}\label{perfomance visualization}
\end{figure*}

In this section, we compare the effectiveness of different account representation methods and classification methods. In the comparison of account representations, we used LINE, control logic representation, CANE, DPSE, and our method to represent contract accounts. In the comparison of classification methods, we compared the effects of SVM and MLP on account embedding classification. The results are listed in Table \ref{result}.

From the table, we have the following observations.

(1) The combination of network structure and operation code information can achieve better results than only using network structure or operation code information. The experimental results show that the method using network structure and operation code information (i.e., CANE, DSPSD) can improve the $F$ value by at least 29\% compared with the method using only structural information (i.e., LINE). In smart Ponzi scheme detection, operation code information is more important than transaction network structure information. Detection methods based only on transaction network structure often have a very low recall value, resulting in a low F-score. For example, according to the performance of DPSE, if only structural information is used for prediction, the recall rate is only 0.32, while the recall rate of using both structure and opcode information is 0.81.

(2) In smart Ponzi scheme detection, the models based on node embedding are better than the models based on feature engineering. One of the reasons is that the recall rate of the feature-based model is lower than that of the node-embedded models. This indicates it is difficult to represent the network structure by several features. The node-embedded model can automatically learn the node representation according to the network topology, which is more accurate than the designed features.

%(2) In smart Ponzi scheme detection, when only structural information is used, the model based on node embedding is better than the model based on feature engineering. When using both structural information and opcode information, the F-score of the node embedding-based model (i.e, CANE, DSPSD) is at least 7\% higher than that of the feature-based model DPSE. It is worth noting that the recall rate of the feature-based model is much lower than the node embedding model. This indicates that it is difficult to represent the network structure with limited number of features. The node-embedded model can automatically learn the node representation according to the network topology, which is more accurate than the designed features.

(3) Compared with support vector machine, MLP has better detection effect. Although the detection performance of MLP and support vector machine both improve with the increase of network information, MLP has better adaptability.

(4) The proposed DSPSD is remarkably better than the baseline methods. When both of the transaction network and opcode information are used, the F-score of DSPSD can reach up to 0.91, which is superior than other methods. Compared with the static node embedding method, DSPSD uses the evolution information of each node, which is of great significance to the detection of smart Ponzi scheme. That indicates that in smart Ponzi scheme detection, we not only need to consider the network structure generated by transaction data, but also the process of generating the transaction network.

%(4) The proposed DSPSD is remarkably better than the baseline methods. When the combination of transaction network and opcode information is used, the F-score of DSPSD can be as high as 0.91, which is far better than the results of other methods. Compared with the static node embedding method, DSPSD uses the evolution information of each node, and this information is of great significance to the smart Ponzi scheme detection.

We used 10-fold cross validation in our experiments, so we have different results on 10 data sets for each experiment. Figure \ref{perfomance visualization} shows the range of values for the results of each method. It can be seen that the result range of DSPSD is smaller than other methods (the corresponding line is shorter), which indicates DSPSD is more robust than the baseline methods.

\begin{figure*}[t]
    \centering
        \subfigure[LINE]{
            \begin{minipage}[t]{0.4\linewidth}
            \centering
            \includegraphics[width=2.2in]{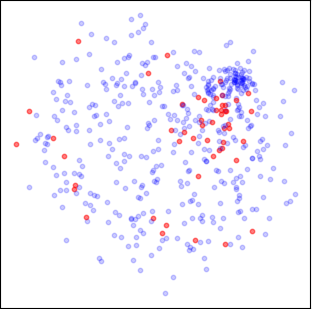}
            \label{line_emb}
            \end{minipage}%
        }% 
          \subfigure[Control logic representation]{
            \begin{minipage}[t]{0.4\linewidth}
            \centering
            \includegraphics[width=2.2in]{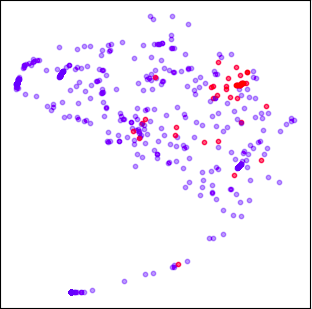}
            \label{cnn_emb}
            \end{minipage}
        }
		
        \subfigure[CANE]{
            \begin{minipage}[t]{0.4\linewidth}
            \centering
            \includegraphics[width=2.2in]{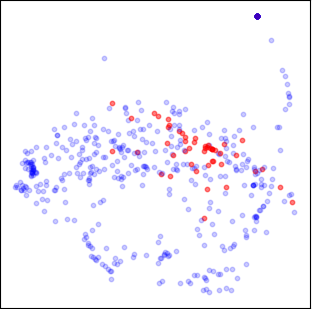}
            \label{cane_emb}
            \end{minipage}%
        }%
        \subfigure[DSPSD]{
            \begin{minipage}[t]{0.4\linewidth}
            \centering
            \includegraphics[width=2.2in]{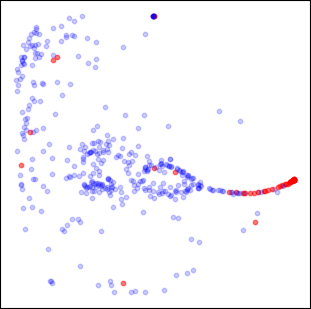}
            \label{dspsd_emb}
            \end{minipage}
        }%
    \centering
    \caption{Embedding Visualization. The red points represent the accounts that implement the smart Ponzi scheme, and the blue points represent the normal accounts.}\label{ponzivis}
\end{figure*}

\subsection{Case Study}
In this part, we have selected two cases for case analysis in the results of false positive and false negative.

\textbf{False positive}:

(1) '0x160fc84c8c5d46561b01d38eb7d44671f3eed4ca': In the data we use, there are only two transaction records related to this account. We guess that the wrong classification is caused by insufficient information.

(2) '0x7753a9d834844cfde5c211ec3912b49f0d8b8e11': There are multiple transaction records with this account in the data we use, but the operation code of this account is missing, which causes our algorithm to generate wrong results.

\textbf{False negative}:

(1) '0xa9fa83d31ff1cfd14b7f9d17f02e48dcfd9cb0cb': The characteristic of transactions related to this account is that the transaction value generated by most accounts is very low, but the transaction value of a few transactions is particularly high. The account is incorrectly classified as an normal account might be because the number of transactions related to the account is similar to that of an normal account, but the transaction value characteristics are different from that of an normal account. In the transaction network, we use the number of transactions instead of the transaction value to define the edge weights. The advantage is that the difference in edge weights will not be too large. Otherwise, the edge with extremely high weight will dominate the representation of the node, while the remaining edges have a very limited influence on the representation of the node. But the disadvantage is that the node representation does not contain information about the transaction value.

%In the data we use, there are 31 transaction records related to the account, 28 of which are less than 0.05 ETH, and the remaining 3 transactions have a transaction value between 1.5 ETH and 10 ETH. The average number of transactions in a normal account is XX, the average transaction value is XX, and the standard deviation of the transaction value is XX. Therefore, the number of transactions in this account is similar to that of a normal account, but the transaction value characteristics are different from that of a normal account. 

%The characteristic of transactions related to this account is that the value of transactions generated with most accounts is very low, but the transaction value of a few transactions is particularly high. The reason why this account was mistakenly classified as a normal account may be that we used the number of transactions instead of the transaction amount when calculating the weights of the edges of the transaction network. The advantage of using the number of transactions is that the difference in edge weights is not too large. Otherwise, the edge with extremely high weight will dominate the representation of the node, while the remaining edges have very limited effect on the representation of the node. But its disadvantage is that it does not contain information about the transaction amount in the node representation.

(2) '0x258d778e4771893758dfd3e7dd1678229320eeb5': The characteristic of transactions related to this account is that the transaction amount is either 1 or 10. We guess that the reason for the wrong classification is also related to the edge weights of the transaction network is defined by the number of transactions instead of the amount.

In summary, false positive cases may be caused by insufficient information. The false negative cases may be caused by the fact that we only used information about the number of transactions and not the value of transactions when constructing the transaction network.
%To sum up, the false negative case may be caused by insufficient information. The false positive case is related to the fact that we only used the number of transactions but not the value of each transaction when we constructed the edge weights of the transaction network. 

\subsection{Embedding Visualization}
\label{sec_exp_emb}

In order to demonstrate the embedding generated by LINE, control logic representation, CANE and DSPSD, we used different methods to generate the embedding of each contract account. The embedding is 200-dimensional, to visualize high-dimensional data, we use t-distributed randocial Neighbor embedding (t-SNE)\cite{DBLP:conf/vissym/RauberFT16} to map each embedding to a two-dimensional vector, and then use 2-D graph display data. The embedding of is shown in Figure \ref{ponzivis}, where the blue points are normal accounts and the red points are smart Ponzi scheme accounts. As can be seen from the figure, the embedding generated by DSPSD is the most classification-friendly. Even from a 2-D graph, it is easy to distinguish the area of the smart Ponzi scheme account vector from the area of the normal contract account vector. On the other hand, the embedding generated by LINE, CANE, and control logic representation are not as good as the embedding generated by DSPSD. Especially for LINE, the area of the smart Ponzi scheme account vector is mixed with the area of the normal contract vector. The embedding generated by CANE is slightly better than the control logic representation because the nodes generated by CANE are more concentrated in a smaller area. This is consistent with the results we obtained in Table \ref{result}.

%It’s not obvious that both LINE, CNN and CANE failed to separate two categories apart clearly. Especially for LINE, two categories are mixed together. We observe that CNN is better than LINE, which only bases on the opcodes information. By introducing opcodes information associated with each account, CANE achieves satisfactory visualizing result among LINE, as the two categories are roughly separated apart from each other. However, there is no clear margin between two categories. As expected, DSPSD can clearly separate two categories apart. The temporal node embeddings of smart Ponzi scheme accounts are different from those of normal accounts. We can see the embeddings of most of the normal accounts are close, while those of the smart Ponzi scheme accounts are not.

\begin{figure}[t]
	\centering
	\includegraphics[width=0.37\textwidth]{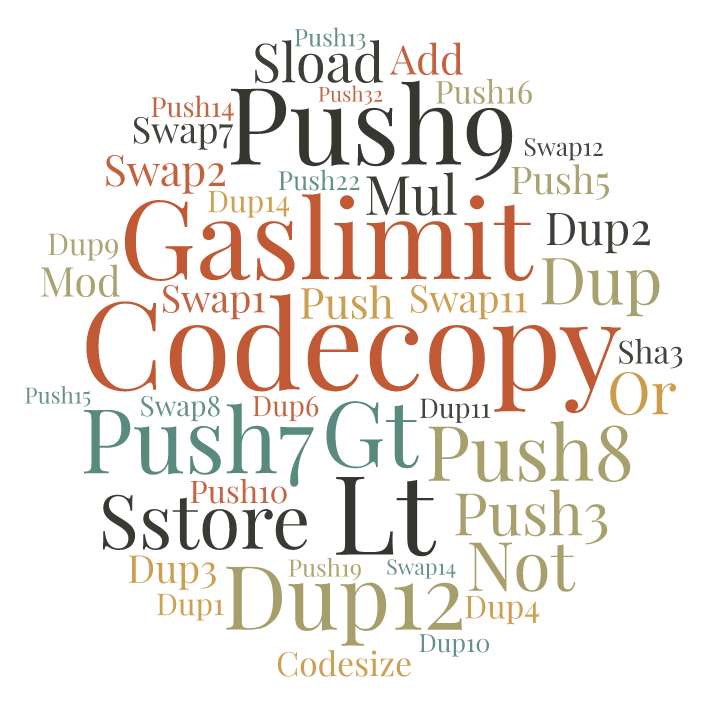}
	\caption{Most important opcodes of Ponzi scheme contract}\label{fig_ponzi_cg}
\end{figure}

\begin{figure}[t]
	\centering
	\includegraphics[width=0.37\textwidth]{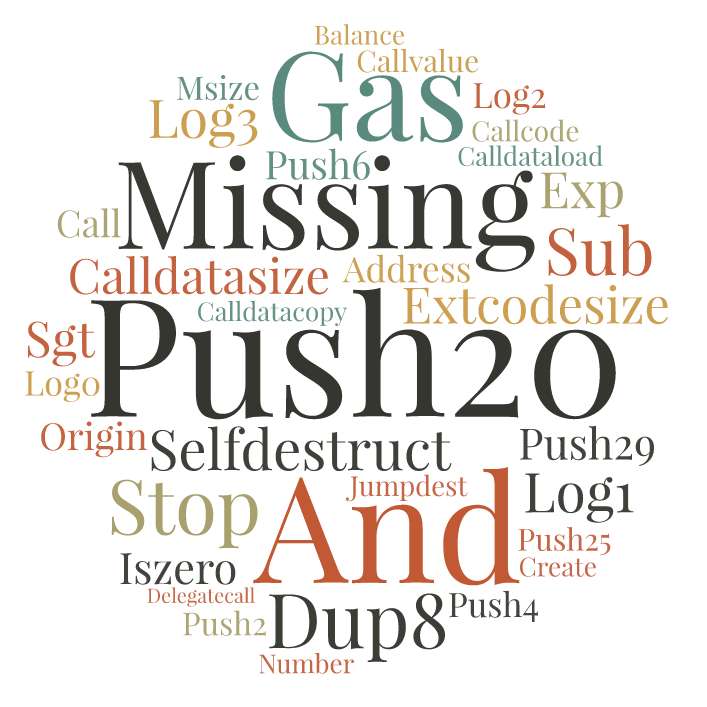}
	\caption{Most important opcodes of normal smart contract}\label{fig_normal_cg}
\end{figure}

\subsection{Opcode Visualization}
\label{sec_exp_op}

Opcodes are important for detecting smart contracts because they reflect the logic of smart contracts. In our experiments, we define important opcodes as those that are strongly related to smart Ponzi scheme contract or normal contract. Generally speaking, an opcode is an important opcode if (1) it is rare and mainly appears in one category, or (2) it frequently appears in one category, but is not common in the other category. 

In our experiment, we use the TF-IDF value to indicate the importance of an opcode. For an opcode $op$, we first compute its IDF value $IDF_{op}$ based on all of the smart contracts. Second, we calculate the TF-IDF value of $op$ in the smart Ponzi scheme contract $TFIDF_{op, p}$ and the TF-IDF value of $op$ in the normal contract $TFIDF_{op, n}$. The final TF-IDF value of $op$ is $max(TFIDF_{op, p}, TFIDF_{op, n})$. Then we select the 80 opcodes with the highest TF-IDF value for display. For one of the 80 opcodes, if its $TFIDF_{op, p}> TFIDF_{op, n}$, we classify it as an opcode related to the smart Ponzi scheme contract and display it in Fig. \ref{fig_ponzi_cg}, otherwise it will be regarded as an opcode related to the normal contract and displayed in Fig. \ref{fig_normal_cg}. The font size is determined by the TF-IDF value of the corresponding opcode.

% we select 80 important opcodes based on their TF-IDF values. Assume $op$ is an opcode in the 80 opcodes, if $TFIDF_{op, p} > TFIDF_{op, n}$, the opcode is a smart Ponzi scheme relavent opcode, otherwise, it is a normal opcode. 

%We show the Ponzi scheme relevant opcodes and normal opcodes in Figure \ref{fig_ponzi_cg} and Figure \ref{fig_normal_cg}, respectively. 
\section{Conclusion and Future Work}
This paper proposes a data-driven smart Ponzi scheme system DSPSD. The system uses historical transaction data of Ethereum as input to construct a dynamic transaction network, and then uses the temporal-point node embedding model we designed to embed the Ethereum account information (including account-related network structural information, network dynamic information, and account attribute information) into a low-dimensional vector, and then the vector is input into a classifier composed of Multi-layer Perceptrons to determine whether the corresponding account has implemented a smart Ponzi scheme. Compared with the traditional feature-based smart Ponzi scheme detection method, DSPSD requires very limited human interaction, which saves a lot of time for feature engineering. In addition, the experimental results show that DSPSD has significantly better performance than the existing smart Ponzi scheme detection method, especially in boosting the recall rate.

This work is the first attempt to implement a data-driven smart Ponzi scheme detection algorithm using dynamic graph embedding technology and has achieved good results, proving that the method is feasible in smart Ponzi scheme detection or related tasks. However, there are still some problems in our work that require further research and improvement. First of all, when we set the edge weights of the transaction network, only the number of transactions data is used and the transaction value is not used. The use of the number of transactions instead of the transaction value is to avoid that the difference of edge weight is too large, which makes the representation of the network node overly dependent on the edge with a large weight. However, this also makes the information of the transaction value not used. How to add transaction value information to the node representation is one of the issues that need to be studied in the future. Second, when constructing the transaction network, we treat all types of accounts on Ethereum as nodes of the same type, ignoring the attribute information of the nodes. Future work can consider how to include the attribute information of the node into the account representation. Using heterogeneous nodes to represent different types of accounts, and then using heterogeneous graph embedding technology to learn the node embedding can be one of the solutions.

\bibliographystyle{IEEEtran}
\bibliography{ref}

\end{document}